\documentclass[runningheads]{llncs}
\usepackage[T1]{fontenc}
\usepackage[misc]{ifsym}
\usepackage{gensymb}
\usepackage{graphicx}
\usepackage{array}
\usepackage{cite}
\usepackage{amsmath,amssymb,amsfonts}
\usepackage{algorithmic}
\usepackage{graphicx}
\usepackage{textcomp}
\usepackage{multirow}
\usepackage{booktabs}
\usepackage{url}
\usepackage{tablefootnote}
\usepackage{stfloats}
\usepackage{color}
\usepackage{verbatim}
\usepackage{bigstrut}
\usepackage{makecell}
\usepackage{caption}
\usepackage[marginal]{footmisc}

\usepackage{float}
\begin{document}
\begin{sloppypar}
\title{Advancing Brain Imaging Analysis Step-by-step via Progressive Self-paced Learning}
\titlerunning{Progressive Self-paced Learning}
%
\author{Yanwu Yang\inst{1,2} 
\and
Hairui Chen\inst{1,2} 
\and
Jiesi Hu\inst{1,2} 
\and
Xutao Guo \inst{1,2} 
\and
Ting Ma \inst{1,2,3}$^{(\textrm{\Letter})}$ 
}

%
\authorrunning{Y. Yang et al.}
%
\institute{
Harbin Institute of Technology at Shenzhen, Shenzhen 518000, China\\ \email{tma@hit.edu.cn}
\\
 \and
Peng Cheng Laboratory, Shenzhen 518066, China \\
\and 
Guangdong Provincial Key Laboratory of Aerospace Communication and Networking Technology, Harbin Institute of Technology (Shenzhen), Shenzhen, China
}
\maketitle              
\footnote{Y. Yang and H. Chen contributed equally to this work.}
\begin{abstract}
Recent advancements in deep learning have shifted the development of brain imaging analysis. However, several challenges remain, such as heterogeneity, individual variations, and the contradiction between the high dimensionality and small size of brain imaging datasets. These issues complicate the learning process, preventing models from capturing intrinsic, meaningful patterns and potentially leading to suboptimal performance due to biases and overfitting.
Curriculum learning (CL) presents a promising solution by organizing training examples from simple to complex, mimicking the human learning process, and potentially fostering the development of more robust and accurate models.
Despite its potential, the inherent limitations posed by small initial training datasets present significant challenges, including overfitting and poor generalization.
In this paper, we introduce the Progressive Self-Paced Distillation (PSPD) framework, employing an adaptive and progressive pacing and distillation mechanism. This allows for dynamic curriculum adjustments based on the states of both past and present models. The past model serves as a teacher, guiding the current model with gradually refined curriculum knowledge and helping prevent the loss of previously acquired knowledge. We validate PSPD's efficacy and adaptability across various convolutional neural networks using the Alzheimer's Disease Neuroimaging Initiative (ADNI) dataset, underscoring its superiority in enhancing model performance and generalization capabilities. The source code for this approach will be released at \url{https://github.com/Hrychen7/PSPD}.

\keywords{curriculum learning  \and brain imaging \and knowledge distillation.}
\end{abstract}
\section{Introduction}
Recently, advanced artificial intelligence technologies such as deep learning have significantly shifted the paradigm of brain imaging analysis \cite{shen2017deep,zhang2020survey,lemm2011introduction}. Prominent achievements of prior works demonstrate that deep learning approaches have become state-of-the-art solutions for a variety of brain imaging analysis problems, such as brain disease diagnosis \cite{li2014deep,yangMappingMultimodalBrain2023,yang2024brainmass}, behavioral phenotype prediction \cite{gong2021phenotype,yang2023deep}, and brain lesion segmentation \cite{bakas2018identifying,henschel2020fastsurfer}.

Nevertheless, there remain great challenges in brain imaging analysis using deep learning approaches. First, brain imaging exhibits heterogeneity, as multiple concurrent pathological processes can cause diverse changes in the brain. Individuals within the same phenotype show varied characteristics and backgrounds, leading to variability in brain imaging patterns \cite{segal2023regional,dujardin2020tau}. Moreover, variations in the quality of brain imaging preprocessing (e.g., registration) might introduce further noise, potentially hindering the training process. Finally, while brain imaging data are inherently high-dimensional, they often come in limited sizes. This discrepancy exacerbates the difficulty of the task. Learning from insufficient samples may result in suboptimal performance due to potential biases and overfitting \cite{benkarim2022population,yang2023creg,yang2022regularizing}. These general sources of variation may hinder predictive models from accurately learning patterns, thereby diminishing their effectiveness.

In this regard, curriculum learning (CL) \cite{bengio2009curriculum,wang2021survey} offers an efficient approach to model training, wherein examples are not presented randomly but are organized in a meaningful sequence. This approach is inspired by the human learning process, which progressively integrates samples from 'easy' to 'hard' during training. Such a paradigm facilitates improved learning by guiding the model to grasp basic (easy) concepts earlier and more advanced (hard) concepts later, thus fostering the development of a more robust model. The effectiveness of this strategy has been validated across various tasks, demonstrating significant performance improvements \cite{li2023dynamic,peng2021self}. However, within the CL framework, training datasets start small and gradually expand to encompass the entire dataset. Given the limited data samples in brain imaging, there is a significant challenge in learning from such insufficient samples, especially in the initial phase, leading to severe overfitting issues and impairing the model’s generalization capabilities. Moreover, due to variations in the model and curriculum during training, there is a risk that the model might concentrate more on the current state and neglect previously acquired valuable knowledge, leading to decreased performance.

To address the above-mentioned issues, we aim to propose an efficient CL framework to improve the performance and generalization capabilities for brain imaging analysis. Our contributions are as follows: 
\textbf{(1)} We introduce the Progressive Self-Paced Distillation (PSPD) framework to regularize model training through a progressive and adaptive pacing mechanism. By applying self-knowledge distillation, PSPD progressively refines the model with dynamically paced knowledge to improve generalization and prevent forgetting prior knowledge during the model's training.
\textbf{(2)} Moreover, unlike most existing methods that leverage fixed, and static curriculum settings, we introduce a decoupled paced curriculum setting. We employ paced curriculum learning (PCL) and paced curriculum distillation (PCD) to provide a curriculum customized to the current and past models' states, respectively. This enables PSPD to create a more effective and dynamically regulated training curriculum, leveraging insights from both the past and present.
\textbf{(3)} We validate the efficacy and adaptability of PSPD through experiments conducted on three well-established convolutional neural networks, utilizing the Alzheimer's Disease Neuroimaging Initiative (ADNI) dataset.
Our evaluations demonstrate the superiority and adaptability of PSPD in enhancing performance and generalization for brain imaging analysis.


\textbf{Related works:} The advantages of
applying curriculum learning strategies has been proven to improve performance and accelerate the training process in various studies. For instance, Jimenez et al. \cite{jimenez2019medical} investigate the application of curriculum learning in X-ray image classification, demonstrating its significant potential in aiding computer-aided diagnosis.
Nebbia et al. \cite{nebbia2021radiomics} suggest using the radiomics score to modulate the loss of each image for Breast Cancer Diagnosis. \cite{li2023dynamic} proposes to estimate the uncertainty score for medical image classification.
Furthermore, the concept of automated curriculum learning, or self-paced learning, has been introduced to increase training efficiency by adaptively quantifying samples, thereby obviating the need to process the entire training set. Peng et al. \cite{peng2021self} incorporate self-paced learning with contrastive learning for medical image segmentation by measuring the difference between different views of an image as the difficulty score.
Liu \cite{liu2022margin} presents a margin-preserving self-paced contrastive learning approach for unsupervised domain adaptation in segmentation. Similarly, Islam et al. \cite{islam2023paced} combine prediction uncertainty with annotation boundary uncertainty to devise a paced-curriculum learning strategy for segmentation. However, these studies potentially overlook the generalizability of the model when organizing insufficient data samples, especially for small and heterogeneous 3D brain imaging datasets.

\section{Method}
\begin{figure}[t] 
    \centering 
    \includegraphics[width=1.0\textwidth ]{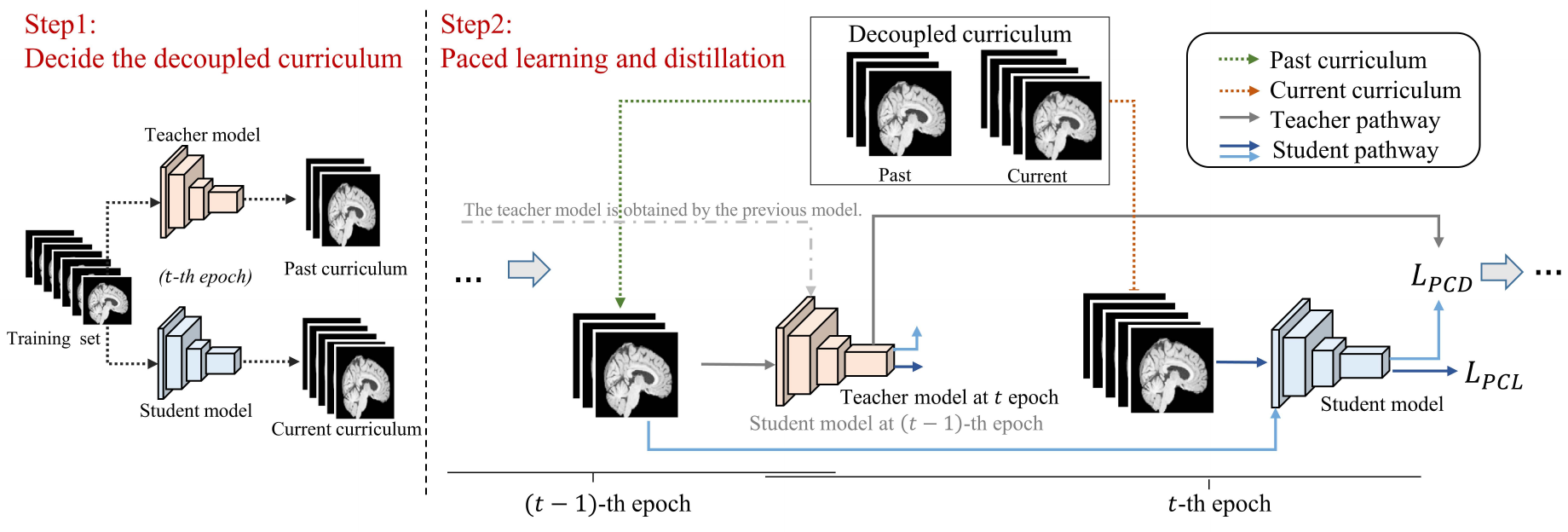} 
    \caption{
    The architecture of the proposed PSPD, where the teacher model is derived from the models trained at previous epochs. We first establish a decoupled curriculum setting that considers both the current and past states of the model, and then train the model based on this setting for learning and distillation.}
    \label{model} 
\end{figure}
As is shown in Fig. \ref{model}, the proposed PSPD is implemented in a self-knowledge distillation framework. We tackle the current model at $t$-th epoch as the student model, and the trained previous model at $(t-1)$-th epoch as the teacher model. In the following sections, we would like firstly to introduce the self-pace strategy, which plays a key role in adaptive curriculum settings. Furthermore, we leverage the self-pace strategy into the progressive self-knowledge distillation framework for a consistent curriculum.

\subsection{Self-paced learning and regularizer}

In the self-paced learning paradigm, the objective function incorporates an importance weight $w_i \in [0,1]$ for the specific loss $l_i$ of each sample $i$. This is obtained by using a self-paced regularization term $R_{\lambda}(w_i)$, which ensures the weights are monotone decreasing with respect to the loss $l_i$ (i.e., harder examples are given less importance) and monotone increasing with respect to the learning pace (i.e., a larger $\lambda$ increases the weights). For each objective function, it is obtained as:
\begin{equation}
    L=\frac{1}{N}\sum_i^N w_i l_i + R_{\lambda}(w_i)
\end{equation}
The pace weights are decided by the regularizer $R_{\lambda}$. We adapt two strategies:

\textbf{Hard}. For the hard training, we gradually reduce the number of discarded hard examples. In the first epoch, only a few easy examples are used for training, and the other examples are discarded. As training continues, more hard examples are retained, and only a small number are discarded. In the last epoch, the proportion of selected samples gradually increases until it encompasses the entire dataset. In this regard, we obtain that:
\begin{equation}
    R^{hard}_\lambda(w_i,l_i)= -\lambda w_i; w_i(l_i, \lambda) = \begin{cases} 1, & \text{if } l_i < \lambda \\ 0, & \text{if } l_i \geq \lambda \end{cases}
\end{equation}

\textbf{Soft}. For the soft training, all the samples are fed into the model training, and each sample is assigned an importance weight. The weights are obtained based on a linear imputation:
\begin{equation}
    R^{soft}_\lambda(w_i,l_i)= \lambda (\frac{1}{2}w_i^2-w_i); w_i(l_i, \lambda) = \begin{cases} 1-\frac{1}{\lambda}l_i, & \text{if } l_i < \lambda \\ 0, & \text{if } l_i \geq \lambda \end{cases}
\end{equation}
In terms of this, we can determine the difficulty level of the samples by $l_i$. The settings of $l_i$ will be introduced in the following sections.

\subsection{Progressive Self-Paced Distillation}
In this study, we leverage the self-knowledge distillation framework to progressively pace the learning progress and refine its curriculum knowledge. In particular, the teacher model is obtained by the past model at $(t-1)$-th epoch, which is dynamically evolved as training proceeds. The loss function is obtained as:
\begin{equation}
    L=\frac{1}{N}\sum_{i=1}^N L_{CE}(y_i,p^S(x_i)) +   \gamma L_{KL}(p^T(x_i), p^S(x_i))
\end{equation}\label{eq1}
where $p^S(x_i)$ and $p^T(x_i)$ are the outputs of the student model and the teacher model respectively. The loss function is a weighted combination of the cross-entropy loss $L_{CE}(\cdot)$ and the KL divergence loss $L_{KL}(\cdot)$ with $\gamma$.

In this paper, the curriculum settings are decided on both the current and the past predictions of the model. The paced curriculum is divided into two folds: based on the current and the past state of the model.

\subsubsection{Paced curriculum learning}
For the current curriculum setting, we propose the paced curriculum learning for the model learning. The paced curriculum learning adaptively measures the difference between the predictions of the current model and the actual labels:
\begin{align}
    &L_{PCL} = \frac{1}{N} \sum_{i=1}^N w_i L_{CE}(p^S(x_i),y_i) + R_{\lambda_w}(w_i, l^w_i) \\
    &l^w_i = L_{CE}(p^S(x_i),y_i)
\end{align}
In this manner, the model is able to learn adaptively filtered samples based on the current condition.
\subsubsection{Paced curriculum distillation}
Paced curriculum distillation utilizes the teacher model to gradually refine the student model and prevent it from forgetting previous useful knowledge. However, since the teacher model may not always impart meaningful knowledge, guiding the student model with two inconsistent targets can be problematic. To address this, we consider the confidence level of the teacher's knowledge. When the teacher provides confident and meaningful knowledge, it can regularize or even calibrate the student model's training. Additionally, if the discrepancy is too significant, the student model will rely solely on the ground truth for learning, without incorporating guidance from the teacher model. Thus, paced knowledge distillation is implemented as:
\begin{align}
    &L_{PCD} = \frac{1}{N} \sum_{i=1}^N \varphi_i L_{KL}(p^T(x_i),p^S(x_i)) + R_{\lambda_\varphi}(\varphi_i, l^\varphi_i) \\
    &l^\varphi_i = L_{CE}(p^T(x_i),y_i)
\end{align}

\subsection{Optimization}
The pace parameter $\lambda_w$ and $\lambda_\varphi$ are set in a linear increase during training as $\lambda_w = \lambda_{w,0} + \alpha_w t$ and $\lambda_\varphi = \lambda_{\varphi,0} + \alpha_\varphi t$, where $t$ denotes the number of the current epoch.
Finally, the objective function is obtained by a combination of paced curriculum learning and distillation as:
\begin{equation}
    \begin{split}
    L = 
    &\frac{1}{N}\sum_i^N w_i L_{CE}(p^S(x_i), y_i) + \gamma \varphi_i  L_{KL}(p^T(x_i),p^S(x_i)) \\ 
    &+ R_{\lambda_w}(l^w, w_i) + R_{\lambda_\varphi}(l^\varphi, \varphi_i)
    \end{split}
\end{equation}

\section{Experiments}
\subsection{Dataset and experimental settings}
\textbf{Dataset.} In this study, we use the 3D T1 images from the Alzheimer's Disease Neuroimaging Initiative (ADNI) database \cite{jack2008alzheimer} (\url{http://www.adni-info.org/}) to evaluate the brain imaging analysis. We built our cohort based on 370 patients that are diagnosed with mild cognitive impairments (MCI) at baseline, and 364 healthy controls (HC). 
The sex and age of the HC and MCI groups are matched.

All the images were preprocessed by AC-PC aligns, brain skull stripping, bias field correction, and normalization into the standard MNI space. All images are down-sampled into the standard 2mm$^3$  space and padded into $96\times 112\times 96$.


\textbf{Implementations}.
We employ three backbones, i.e., ResNet-18, ResNet-50, and ResNet-101, to show the adaptability of our proposed PSPD on different backbones. All these models are trained by the Adam optimizer. The initial learning rate is 1e-6 and increases to 1e-4 in 10 warmup epochs. 
The models are trained in 180 epochs. The initial values of the pace parameters $\lambda_{w,0}$ and $\lambda_{\varphi,0}$ are set as 0.6 and 0.8. $\alpha_w$ and $\alpha_\varphi$ are set as 0.006 and 0.003 respectively.
We apply 4-fold cross-validation, where 70\%, 15\%, and 15\% samples were randomly selected for training, validation, and testing. We use the diagnosis accuracy (Acc), sensitivity (Sen), Specificity (Spe), and area under the curve (AUC) for evaluation. In addition, we use the Expected Calibration Error (ECE) and negative log loss (NLL) to measure the confidence of the prediction, indicating the overfitting/generalization of the model. Small values indicate better generalization.

\textbf{Competitive baseline}.
We compare our proposed method with two related CL methods, i.e., CuDFKD \cite{li2023dynamic1}, Dy-KD \cite{lin2023dy}, and two CL methods for medical images, i.e., Dynamic Curriculum Learning via In-Domain Uncertainty (DLCU) \cite{li2023dynamic}, and Medical-based Deep Curriculum Learning (MDCL) \cite{jimenez2019medical}. We also compare with the baseline backbone without curriculum learning. For a fair comparison, all these baseline methods are compared by implementing a grid search of the parameters to decide the curriculum settings.

\subsection{Sensitivity analysis and ablation studies}
\begin{figure}[t]
    \centering
    \includegraphics[width=1.0\textwidth ]{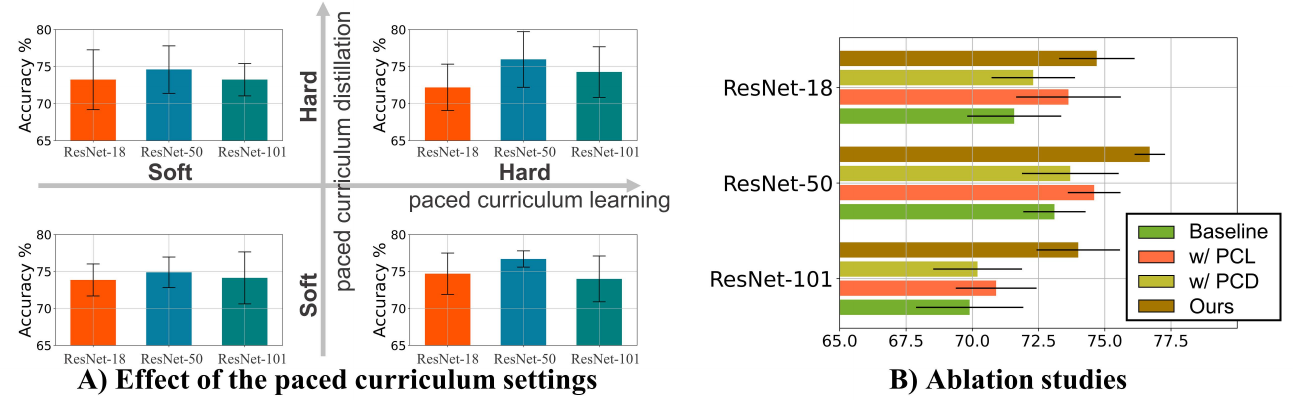}
    \caption{The results of different curriculum approaches and ablation studies.} 
    \label{sen}
\end{figure}

\textbf{Effect of the way of pace curriculum.} We evaluated the way of the paced curriculum settings, i.e., the soft and hard ways. As shown in Fig. \ref{sen} A), the hard and soft training methods for PCL achieve comparable performances. However, for PCD, the soft approach outperforms the hard in most cases. In this regard, we implement the soft training settings for PCD in the following studies. The PCL is implemented by hard training as default.

\textbf{Ablation studies.} To further delineate the contributions of each component, we conducted ablation studies to identify which element plays a more critical role in our framework.
The results are displayed in Fig. \ref{sen} B). Compared to the baseline, both PCL and PCD consistently improved their performance. Notably, PCD plays a more significant role than PCL, contributing to larger improvements. This can be attributed to the self-knowledge distillation paradigm, on the one hand, which serves as a regularization term, penalizing the predictions and refining the model. On the other hand, our paced curriculum settings regularize or even calibrate the predictions with previously learned knowledge.
\begin{figure}[t]
    \centering 
    \includegraphics[width=1.0\textwidth ]{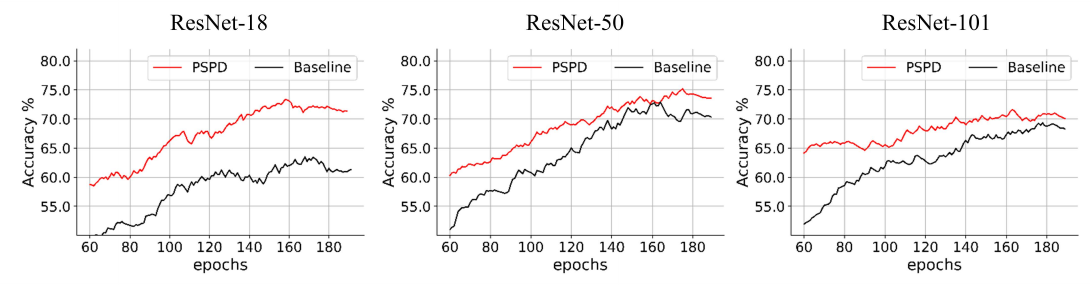}
    \caption{The learning curve of the validation accuracy.} 
    \label{curve} 
\end{figure}
\begin{table}[h]
    \centering
    \setlength{\tabcolsep}{5pt}
    \scriptsize
    \caption{Classification performance on three backbones of six methods with average and standard deviation across folds (Mean-Std). The best results are shown in bold.}
\begin{tabular}{r|c|cccccc}
\hline
\hline
\multicolumn{1}{c|}{Backbone} & Metric & Baseline & CuDFKD  & Dy-kD & DLCU  & MBCL  & PSPD (ours) \bigstrut\\
\hline
\multicolumn{1}{r|}{\multirow{4}[2]{*}{ResNet18}} & ACC $\uparrow$ & 71.6-3.5 & 72.3-3.1 & 73.3-1.2 & 73.0-1.8 & 73.6-6.4 & \textbf{74.7-2.8} \\
      & SEN $\uparrow$ & 77.0-10.4 & 81.8-10.5 & \textbf{83.1-8.2*} & 80.1-2.6 & 78.4-13.1 & 78.4-5.1 \\
      & SPE $\uparrow$ & 68.9-6.6 & 69.4-4.0 & 68.5-2.6 & 68.9-2.0 & 68.8-5.3 & \textbf{70.8-4.6} \\
      & AUC $\uparrow$ & 78.2-2.7 & 78.5-1.9 & 79.1-1.3 & 78.1-3.2 & 79.2-2.5 & \textbf{80.7-0.8}\\
\hline
\multicolumn{1}{r|}{\multirow{4}[2]{*}{ResNet50}} & ACC $\uparrow$ & 73.1-2.3 & 73.7-3.6 & 75.34-5.73 & 75.3-3.4 & 73.0-1.4 & \textbf{76.7-1.1*} \bigstrut[t]\\
      & SEN $\uparrow$ & 76.6-6.7 & 77.5-9.9 & 83.04-5.7* & 84.5-9.4* & 75.0-10.7 & \textbf{83.8-2.2*} \\
      & SPE $\uparrow$ & 69.4-2.3 & 69.7-4.7 & 72.4-10.11 & \textbf{77.8-5.6*} & 72.9-8.4 & 69.4-2.3 \\
      & AUC $\uparrow$ & 78.3-2.2 & 79.9-2.6 & 80.7-1.44 & 80.4-2.5 & 78.2-2.1 & \textbf{81.1-1.0} \bigstrut[b]\\
\hline
\multicolumn{1}{r|}{\multirow{4}[2]{*}{ResNet101}} & ACC $\uparrow$ & 69.9-4.0 & 70.2-3.3 & 72.6-3.87 & 71.6-3.4 & 71.8-3.3 & \textbf{74.0-3.1*} \bigstrut[t]\\
      & SEN $\uparrow$ & 73.0-9.6 & 79.7-10.9 & 84.5-7.73* & 73.0-4.3 & 83.1-6.1* & \textbf{87.8-5.6*} \\
      & SPE $\uparrow$ & 66.7-3.9 & 66.7-5.5 & 65.7-3.46 & 70.1-7.7 & \textbf{76.4-4.2} & 75.0-3.0 \\
      & AUC $\uparrow$ & 77.3-3.1 & 76.2-5.9 & 79.5-3.0 & 77.3-3.8 & 78.4-3.7 & \textbf{80.8-3.5} \bigstrut[b]\\
\hline
\hline
\end{tabular}%
\label{tab1}%
\begin{flushleft}
\scriptsize{
* indicates significant improvements ($p<0.05$)}
\end{flushleft}
\end{table}
\subsection{Evaluation on the classification performance}
Table \ref{tab1} presents the classification results for the baseline, four competitive methods, and our PSPD across three ResNet architectures. It is evident that all curriculum learning methods, including ours, enhance performance relative to the baseline. Notably, ResNet-50 outperforms ResNet-18 and ResNet-101 in most scenarios, attributed to ResNet-101's susceptibility to overfitting due to its deeper architecture. PSPD consistently achieves superior accuracy and AUC scores on all three architectures, with improvements over the baseline by 3.1\%, 3.6\%, and 4.1\% in accuracy, respectively. Furthermore, the learning curves depicted in Fig. \ref{curve} reveal smaller fluctuations for PSPD compared to the baseline, indicating a more stable and robust training process.

\subsection{Evaluation on generalization performance}
We present the expected calibration error rate and the negative log loss in Table \ref{tab2}. These two metrics evaluate the quality of predictive probabilities and confidence estimation serving as indicators of overfitting. The results demonstrate that our PSPD records the lowest error rates among all compared methods. Furthermore, while the DLCU achieves the second best in most cases as shown in Table \ref{tab1}, its confidence estimation is worse than the baseline, suggesting reduced robustness and potential generalization issues. Overall, PSPD not only enhances performance but also improves generalization and reduces overfitting.
\begin{table}[h]
    \centering
    \scriptsize
    \setlength{\tabcolsep}{5pt}
    \caption{ECE (\%) and NLL results on three backbones of six methods. The results are shown with mean and standard deviation across folds (Mean±Std).}
\begin{tabular}{r|c|cccccc}
\hline
\hline
\multicolumn{1}{c|}{Backbone} & Metric & Baseline & CuDFKD  & Dy-kD & DLCU  & MBCL  & PSPD (ours) \bigstrut\\
\hline
\multicolumn{1}{r|}{\multirow{2}[2]{*}{ResNet18}} & ECE $\downarrow$ & 11.38-2.60 & 14.43-3.01 & 13.18-4.30 & 18.71-7.61 & 10.79-3.49 & \textbf{9.50-2.53} \bigstrut[t]\\
      & NLL $\downarrow$ & 0.57-0.03 & 0.63-0.02 & 0.67-0.08 & 1.29-0.06 & \textbf{0.54-0.03} & \textbf{0.54-0.02} \bigstrut[b]\\
\hline
\multicolumn{1}{r|}{\multirow{2}[2]{*}{ResNet50}} & ECE $\downarrow$ & 11.02-3.48 & 9.55-0.52 & 12.28-2.93 & 18.01-6.06 & 12.25-3.34 & \textbf{10.26-3.42} \bigstrut[t]\\
      & NLL $\downarrow$ & 0.59-0.03 & 0.56-0.03 & 0.59-0.04 & 0.94-0.12 & 0.64-0.07 & \textbf{0.53-0.02*} \bigstrut[b]\\
\hline
\multicolumn{1}{r|}{\multirow{2}[2]{*}{ResNet01}} & ECE $\downarrow$ & 11.40-2.37 & 9.21-1.82 & 8.84-2.76 & 21.17-4.23 & 9.63-1.98 & \textbf{8.55-2.88*} \bigstrut[t]\\
      & NLL $\downarrow$ & 0.61-0.02 & 0.59-0.05 & 0.57-0.02 & 0.96-0.14 & 0.59-0.03 & \textbf{0.56-0.08*} \bigstrut[b]\\
\hline
\hline
\end{tabular}%
    \label{tab2}%
\begin{flushleft}
  \scriptsize{
    * indicates significant improvements ($p<0.05$)}
  \end{flushleft}
  \end{table}%

\section{Conclusion}
In this paper, we introduced the progressive self-paced distillation framework, which incorporates a progressive and adaptive pacing mechanism along with self-knowledge distillation to dynamically refine curriculum knowledge, enhance generalization, and prevent the loss of previously acquired knowledge. Our framework stands out by offering decoupled paced curriculum settings, adapting the curriculum to the current and past model states, thus ensuring a more effective and dynamically regulated training process.
Through experiments on several architectures over the ADNI dataset, we have demonstrated PSPD's superiority in improving performance as well as adaptability and generalization capabilities in brain imaging analysis, underscoring its potential for practical use.



\begin{credits}
\subsubsection{\ackname}
The neuroimaging datasets used in this study were supported by the Alzheimer’s Disease Neuroimaging Initiative (ADNI). This work was supported in part by the National Natural Science Foundation of China under Grant 62276081, and in part by the Major Key Project of Peng Cheng Laboratory under Grant PCL2023A09.

\subsubsection{\discintname}
The authors declare that they have no known competing financial interests or personal relationships that could have appeared to influence the work reported in this paper.

\end{credits}

\bibliographystyle{splncs04}
\bibliography{paper-0173}

\begin{thebibliography}{10}
\providecommand{\url}[1]{\texttt{#1}}
\providecommand{\urlprefix}{URL }
\providecommand{\doi}[1]{https://doi.org/#1}

\bibitem{bakas2018identifying}
Bakas, S., Reyes, M., Jakab, A., Bauer, S., Rempfler, M., Crimi, A., Shinohara, R.T., Berger, C., Ha, S.M., Rozycki, M., et~al.: Identifying the best machine learning algorithms for brain tumor segmentation, progression assessment, and overall survival prediction in the brats challenge. arXiv preprint arXiv:1811.02629  (2018)

\bibitem{bengio2009curriculum}
Bengio, Y., Louradour, J., Collobert, R., Weston, J.: Curriculum learning. In: Proceedings of the 26th annual international conference on machine learning. pp. 41--48 (2009)

\bibitem{benkarim2022population}
Benkarim, O., Paquola, C., Park, B.y., Kebets, V., Hong, S.J., Vos~de Wael, R., Zhang, S., Yeo, B.T., Eickenberg, M., Ge, T., et~al.: Population heterogeneity in clinical cohorts affects the predictive accuracy of brain imaging. PLoS biology  \textbf{20}(4),  e3001627 (2022)

\bibitem{dujardin2020tau}
Dujardin, S., Commins, C., Lathuiliere, A., Beerepoot, P., Fernandes, A.R., Kamath, T.V., De~Los~Santos, M.B., Klickstein, N., Corjuc, D.L., Corjuc, B.T., et~al.: Tau molecular diversity contributes to clinical heterogeneity in alzheimer’s disease. Nature medicine  \textbf{26}(8),  1256--1263 (2020)

\bibitem{gong2021phenotype}
Gong, W., Beckmann, C.F., Smith, S.M.: Phenotype discovery from population brain imaging. Medical image analysis  \textbf{71},  102050 (2021)

\bibitem{henschel2020fastsurfer}
Henschel, L., Conjeti, S., Estrada, S., Diers, K., Fischl, B., Reuter, M.: Fastsurfer-a fast and accurate deep learning based neuroimaging pipeline. NeuroImage  \textbf{219},  117012 (2020)

\bibitem{islam2023paced}
Islam, M., Seenivasan, L., Sharan, S., Viekash, V., Gupta, B., Glocker, B., Ren, H.: Paced-curriculum distillation with prediction and label uncertainty for image segmentation. International Journal of Computer Assisted Radiology and Surgery pp.~1--9 (2023)

\bibitem{jack2008alzheimer}
Jack~Jr, C.R., Bernstein, M.A., Fox, N.C., Thompson, P., Alexander, G., Harvey, D., Borowski, B., Britson, P.J., L.~Whitwell, J., Ward, C., et~al.: The alzheimer's disease neuroimaging initiative (adni): Mri methods. Journal of Magnetic Resonance Imaging: An Official Journal of the International Society for Magnetic Resonance in Medicine  \textbf{27}(4),  685--691 (2008)

\bibitem{jimenez2019medical}
Jim{\'e}nez-S{\'a}nchez, A., Mateus, D., Kirchhoff, S., Kirchhoff, C., Biberthaler, P., Navab, N., Gonz{\'a}lez~Ballester, M.A., Piella, G.: Medical-based deep curriculum learning for improved fracture classification. In: Medical Image Computing and Computer Assisted Intervention--MICCAI 2019: 22nd International Conference, Shenzhen, China, October 13--17, 2019, Proceedings, Part VI 22. pp. 694--702. Springer (2019)

\bibitem{lemm2011introduction}
Lemm, S., Blankertz, B., Dickhaus, T., M{\"u}ller, K.R.: Introduction to machine learning for brain imaging. Neuroimage  \textbf{56}(2),  387--399 (2011)

\bibitem{li2023dynamic}
Li, C., Li, M., Peng, C., Lovell, B.C.: Dynamic curriculum learning via in-domain uncertainty for medical image classification. In: International Conference on Medical Image Computing and Computer-Assisted Intervention. pp. 747--757. Springer (2023)

\bibitem{li2023dynamic1}
Li, J., Zhou, S., Li, L., Wang, H., Bu, J., Yu, Z.: Dynamic data-free knowledge distillation by easy-to-hard learning strategy. Information Sciences  \textbf{642},  119202 (2023)

\bibitem{li2014deep}
Li, R., Zhang, W., Suk, H.I., Wang, L., Li, J., Shen, D., Ji, S.: Deep learning based imaging data completion for improved brain disease diagnosis. In: Medical Image Computing and Computer-Assisted Intervention--MICCAI 2014: 17th International Conference, Boston, MA, USA, September 14-18, 2014, Proceedings, Part III 17. pp. 305--312. Springer (2014)

\bibitem{lin2023dy}
Lin, C., Jiang, N., Tang, J., Huang, X., Wu, W.: Dy-kd: Dynamic knowledge distillation for reduced easy examples. In: International Conference on Neural Information Processing. pp. 223--234. Springer (2023)

\bibitem{liu2022margin}
Liu, Z., Zhu, Z., Zheng, S., Liu, Y., Zhou, J., Zhao, Y.: Margin preserving self-paced contrastive learning towards domain adaptation for medical image segmentation. IEEE Journal of Biomedical and Health Informatics  \textbf{26}(2),  638--647 (2022)

\bibitem{nebbia2021radiomics}
Nebbia, G., Dadsetan, S., Arefan, D., Zuley, M.L., Sumkin, J.H., Huang, H., Wu, S.: Radiomics-informed deep curriculum learning for breast cancer diagnosis. In: Medical Image Computing and Computer Assisted Intervention--MICCAI 2021: 24th International Conference, Strasbourg, France, September 27--October 1, 2021, Proceedings, Part V 24. pp. 634--643. Springer (2021)

\bibitem{peng2021self}
Peng, J., Wang, P., Desrosiers, C., Pedersoli, M.: Self-paced contrastive learning for semi-supervised medical image segmentation with meta-labels. Advances in Neural Information Processing Systems  \textbf{34},  16686--16699 (2021)

\bibitem{segal2023regional}
Segal, A., Parkes, L., Aquino, K., Kia, S.M., Wolfers, T., Franke, B., Hoogman, M., Beckmann, C.F., Westlye, L.T., Andreassen, O.A., et~al.: Regional, circuit and network heterogeneity of brain abnormalities in psychiatric disorders. Nature Neuroscience  \textbf{26}(9),  1613--1629 (2023)

\bibitem{shen2017deep}
Shen, D., Wu, G., Suk, H.I.: Deep learning in medical image analysis. Annual review of biomedical engineering  \textbf{19},  221--248 (2017)

\bibitem{wang2021survey}
Wang, X., Chen, Y., Zhu, W.: A survey on curriculum learning. IEEE Transactions on Pattern Analysis and Machine Intelligence  \textbf{44}(9),  4555--4576 (2021)

\bibitem{yang2023creg}
Yang, Y., Guo, X., Ye, C., Xiang, Y., Ma, T.: Creg-kd: Model refinement via confidence regularized knowledge distillation for brain imaging. Medical Image Analysis  \textbf{89},  102916 (2023)

\bibitem{yang2022regularizing}
Yang, Y., Xutao, G., Ye, C., Xiang, Y., Ma, T.: Regularizing brain age prediction via gated knowledge distillation. In: International Conference on Medical Imaging with Deep Learning. pp. 1430--1443. PMLR (2022)

\bibitem{yangMappingMultimodalBrain2023}
Yang, Y., Ye, C., Guo, X., Wu, T., Xiang, Y., Ma, T.: Mapping multi-modal brain connectome for brain disorder diagnosis via cross-modal mutual learning \url{https://ieeexplore.ieee.org/abstract/document/10182318/}, publisher: {IEEE}

\bibitem{yang2023deep}
Yang, Y., Ye, C., Ma, T.: A deep connectome learning network using graph convolution for connectome-disease association study. Neural Networks  \textbf{164},  91--104 (2023)

\bibitem{yang2024brainmass}
Yang, Y., Ye, C., Su, G., Zhang, Z., Chang, Z., Chen, H., Chan, P., Yu, Y., Ma, T.: Brainmass: Advancing brain network analysis for diagnosis with large-scale self-supervised learning. IEEE Transactions on Medical Imaging  (2024)

\bibitem{zhang2020survey}
Zhang, L., Wang, M., Liu, M., Zhang, D.: A survey on deep learning for neuroimaging-based brain disorder analysis. Frontiers in neuroscience  \textbf{14}, ~779 (2020)

\end{thebibliography}


\begin{thebibliography}{10}
\providecommand{\url}[1]{\texttt{#1}}
\providecommand{\urlprefix}{URL }
\providecommand{\doi}[1]{https://doi.org/#1}

\bibitem{antonelli2022medical}
Antonelli, M., Reinke, A., Bakas, S., Farahani, K., Kopp-Schneider, A., Landman, B.A., Litjens, G., Menze, B., Ronneberger, O., Summers, R.M., et~al.: The medical segmentation decathlon. Nature communications  \textbf{13}(1), ~4128 (2022)

\bibitem{bakas2018identifying}
Bakas, S., Reyes, M., Jakab, A., Bauer, S., Rempfler, M., Crimi, A., Shinohara, R.T., Berger, C., Ha, S.M., Rozycki, M., et~al.: Identifying the best machine learning algorithms for brain tumor segmentation, progression assessment, and overall survival prediction in the brats challenge. arXiv preprint arXiv:1811.02629  (2018)

\bibitem{benkarim2022population}
Benkarim, O., Paquola, C., Park, B.y., Kebets, V., Hong, S.J., Vos~de Wael, R., Zhang, S., Yeo, B.T., Eickenberg, M., Ge, T., et~al.: Population heterogeneity in clinical cohorts affects the predictive accuracy of brain imaging. PLoS biology  \textbf{20}(4),  e3001627 (2022)

\bibitem{bernard2018deep}
Bernard, O., Lalande, A., Zotti, C., Cervenansky, F., Yang, X., Heng, P.A., Cetin, I., Lekadir, K., Camara, O., Ballester, M.A.G., et~al.: Deep learning techniques for automatic mri cardiac multi-structures segmentation and diagnosis: is the problem solved? IEEE transactions on medical imaging  \textbf{37}(11),  2514--2525 (2018)

\bibitem{dujardin2020tau}
Dujardin, S., Commins, C., Lathuiliere, A., Beerepoot, P., Fernandes, A.R., Kamath, T.V., De~Los~Santos, M.B., Klickstein, N., Corjuc, D.L., Corjuc, B.T., et~al.: Tau molecular diversity contributes to clinical heterogeneity in alzheimer’s disease. Nature medicine  \textbf{26}(8),  1256--1263 (2020)

\bibitem{gao2022joint}
Gao, S., Zhou, H., Gao, Y., Zhuang, X.: Joint modeling of image and label statistics for enhancing model generalizability of medical image segmentation. In: International Conference on Medical Image Computing and Computer-Assisted Intervention. pp. 360--369. Springer (2022)

\bibitem{gong2021phenotype}
Gong, W., Beckmann, C.F., Smith, S.M.: Phenotype discovery from population brain imaging. Medical image analysis  \textbf{71},  102050 (2021)

\bibitem{henschel2020fastsurfer}
Henschel, L., Conjeti, S., Estrada, S., Diers, K., Fischl, B., Reuter, M.: Fastsurfer-a fast and accurate deep learning based neuroimaging pipeline. NeuroImage  \textbf{219},  117012 (2020)

\bibitem{islam2023paced}
Islam, M., Seenivasan, L., Sharan, S., Viekash, V., Gupta, B., Glocker, B., Ren, H.: Paced-curriculum distillation with prediction and label uncertainty for image segmentation. International Journal of Computer Assisted Radiology and Surgery pp.~1--9 (2023)

\bibitem{jimenez2019medical}
Jim{\'e}nez-S{\'a}nchez, A., Mateus, D., Kirchhoff, S., Kirchhoff, C., Biberthaler, P., Navab, N., Gonz{\'a}lez~Ballester, M.A., Piella, G.: Medical-based deep curriculum learning for improved fracture classification. In: Medical Image Computing and Computer Assisted Intervention--MICCAI 2019: 22nd International Conference, Shenzhen, China, October 13--17, 2019, Proceedings, Part VI 22. pp. 694--702. Springer (2019)

\bibitem{lemm2011introduction}
Lemm, S., Blankertz, B., Dickhaus, T., M{\"u}ller, K.R.: Introduction to machine learning for brain imaging. Neuroimage  \textbf{56}(2),  387--399 (2011)

\bibitem{li2023dynamic}
Li, C., Li, M., Peng, C., Lovell, B.C.: Dynamic curriculum learning via in-domain uncertainty for medical image classification. In: International Conference on Medical Image Computing and Computer-Assisted Intervention. pp. 747--757. Springer (2023)

\bibitem{li2014deep}
Li, R., Zhang, W., Suk, H.I., Wang, L., Li, J., Shen, D., Ji, S.: Deep learning based imaging data completion for improved brain disease diagnosis. In: Medical Image Computing and Computer-Assisted Intervention--MICCAI 2014: 17th International Conference, Boston, MA, USA, September 14-18, 2014, Proceedings, Part III 17. pp. 305--312. Springer (2014)

\bibitem{liu2022margin}
Liu, Z., Zhu, Z., Zheng, S., Liu, Y., Zhou, J., Zhao, Y.: Margin preserving self-paced contrastive learning towards domain adaptation for medical image segmentation. IEEE Journal of Biomedical and Health Informatics  \textbf{26}(2),  638--647 (2022)

\bibitem{menze2014multimodal}
Menze, B.H., Jakab, A., Bauer, S., Kalpathy-Cramer, J., Farahani, K., Kirby, J., Burren, Y., Porz, N., Slotboom, J., Wiest, R., et~al.: The multimodal brain tumor image segmentation benchmark (brats). IEEE transactions on medical imaging  \textbf{34}(10),  1993--2024 (2014)

\bibitem{nebbia2021radiomics}
Nebbia, G., Dadsetan, S., Arefan, D., Zuley, M.L., Sumkin, J.H., Huang, H., Wu, S.: Radiomics-informed deep curriculum learning for breast cancer diagnosis. In: Medical Image Computing and Computer Assisted Intervention--MICCAI 2021: 24th International Conference, Strasbourg, France, September 27--October 1, 2021, Proceedings, Part V 24. pp. 634--643. Springer (2021)

\bibitem{peng2021self}
Peng, J., Wang, P., Desrosiers, C., Pedersoli, M.: Self-paced contrastive learning for semi-supervised medical image segmentation with meta-labels. Advances in Neural Information Processing Systems  \textbf{34},  16686--16699 (2021)

\bibitem{segal2023regional}
Segal, A., Parkes, L., Aquino, K., Kia, S.M., Wolfers, T., Franke, B., Hoogman, M., Beckmann, C.F., Westlye, L.T., Andreassen, O.A., et~al.: Regional, circuit and network heterogeneity of brain abnormalities in psychiatric disorders. Nature Neuroscience  \textbf{26}(9),  1613--1629 (2023)

\bibitem{shen2017deep}
Shen, D., Wu, G., Suk, H.I.: Deep learning in medical image analysis. Annual review of biomedical engineering  \textbf{19},  221--248 (2017)

\bibitem{yang2023creg}
Yang, Y., Guo, X., Ye, C., Xiang, Y., Ma, T.: Creg-kd: Model refinement via confidence regularized knowledge distillation for brain imaging. Medical Image Analysis  \textbf{89},  102916 (2023)

\bibitem{yangMappingMultimodalBrain2023}
Yang, Y., Ye, C., Guo, X., Wu, T., Xiang, Y., Ma, T.: Mapping multi-modal brain connectome for brain disorder diagnosis via cross-modal mutual learning \url{https://ieeexplore.ieee.org/abstract/document/10182318/}, publisher: {IEEE}

\bibitem{yang2023deep}
Yang, Y., Ye, C., Ma, T.: A deep connectome learning network using graph convolution for connectome-disease association study. Neural Networks  \textbf{164},  91--104 (2023)

\bibitem{zhang2020survey}
Zhang, L., Wang, M., Liu, M., Zhang, D.: A survey on deep learning for neuroimaging-based brain disorder analysis. Frontiers in neuroscience  \textbf{14}, ~779 (2020)

\end{thebibliography}
\end{sloppypar}

\end{document}


\begin{sloppypar}
\title{Appendix of \\ Advancing Brain Imaging Analysis Step-by-step via Progressive Self-paced Learning}
%
%
\author{Anonymous}
\authorrunning{F. Author et al.}
%
\institute{Anonymous}
\maketitle              
%

%
%

\section{Proof}
In PSPD, the paced curriculum is divided into two folds: based on the current and the past state of the model. For each term, it can be obtained as follows:
\begin{equation}
    L=\frac{1}{N}\sum_i^N w_i l_i + R_{\lambda}(w_i)
\end{equation}

Through the concurrent optimization of the model parameters denoted as $\theta$ and the importance weights represented by $w$ utilizing an Alternative Optimization Strategy (AOS) with a gradually increasing age parameter, the capability to automatically integrate additional samples is augmented. Moreover, given fixed model parameters $\theta$ the optimal problem becomes 
\begin{equation}
    \min_{w_{i}\in[0,1]} w_{i}l_{i} +R_{\lambda}(w_i)
\end{equation}

The optimal solutions for the hard and soft strategies are provided by the following propositions.

\textbf{Proposition 1}. The definitions of $R^{hard}_\lambda(w_i,l_i)$ and the closed-form solutions $w^{hard}_i(l_i,\lambda)$ are given by
\begin{equation}
    R^{hard}_\lambda(w_i,l_i)= -\lambda w_i; w^{hard}_i(l_i, \lambda) = \begin{cases} 1, & \text{if } l_i < \lambda \\ 0, & \text{if } l_i \geq \lambda \end{cases}
\label{1}
\end{equation}
Proof. For the hard regularizer $R^{hard}_\lambda(w_i,l_i)$, the optimization problem becomes
\begin{equation}
\min_{w_{i}\in[0,1]} w_{i}l_{i} - \lambda w_{i} = (l_{i} - \lambda)w_{i}
\end{equation}
If $l_i<\lambda$, the minimum is achieved when $w_i=1$. Conversely, if $l_i\geq\lambda$, the minimum is achieved when $w_i=0$. Therefore, the optimization result precisely corresponds to the threshold defined in Equation \ref{1}.

\textbf{Proposition 2}.The definitions of $R^{soft}_\lambda(w_i,l_i)$ and the closed-form solutions $w^{soft}_i(l_i,\lambda)$ are given by
\begin{equation}
   R^{soft}_\lambda(w_i,l_i)= \lambda (\frac{1}{2}w_i^2-w_i); w^{soft}_i(l_i, \lambda) = \begin{cases} 1-\frac{1}{\lambda}w_i, & \text{if } l_i < \lambda \\ 0, & \text{if } l_i \geq \lambda \end{cases}
\label{2}
\end{equation}
Proof. For the soft regularizer $R^{soft}_\lambda(w_i,l_i)$, the optimization problem becomes
\begin{equation}
\min_{w_{i}\in[0,1]} w_{i}l_{i} - \lambda (w_i-\frac{1}{2}w_i^2) = \frac{\lambda}{2}w_i^2 -(\lambda-l_i)w_i
\end{equation}
If $l_i\geq\lambda$, the minimum is achieved when $w_i=0$. However, if $l_i<\lambda$, the optimum is determined by differentiating the function with respect to $w_i$ and setting the result to zero, giving
\begin{equation}
w_{i}=1-\frac{1}{\lambda}l_i
\end{equation}
The optimization result is precisely the threshold obtained by Equ.\ref{2}

\section{Pseudo-Code}
\begin{algorithm}
\caption{Progressive Self-paced Learning}\label{algorithm}
\KwIn{ Training Dataset $D_{train}=\{(x_i,y_i),...\}$, Student model $S_\theta$, Teacher model $T_\theta$, $\lambda_\varphi$-scheduler $\lambda_\varphi(t)$, $\lambda_w$-scheduler $\lambda_w(t)$}
\KwOut{Parameters $\theta$ of student model}
\For{$Epoch = 1,...,T$}
{Update $\lambda_\varphi$ and $\lambda_w$ by $\lambda_\varphi \leftarrow \lambda_\varphi(t)$ and $\lambda_w \leftarrow \lambda_w(t)$\;
Determine the current curriculum by  $S_\theta$ on dataset $D$ using
$l^{w}_i=L_{CE}(p^S(x_i),y_i)$\;

Compute the current curriculum weight $w\leftarrow w(l^w_i,\lambda)$\;
\eIf{$Epoch = 1$}
{Train the student model $S_\theta$ by the $L_{PCL}$ with weight $w$ }
{Determine the past curriculum by $S_\theta$ on dataset $D$ using
$l^{\varphi}_i=L_{CE}(p^T(x_i),y_i)$\;
Compute the past curriculum weight $\varphi\leftarrow\ \varphi(l^\varphi,\lambda)$\;
Train the student model $S_\theta$ by  $L_{PCL}$ and $L_{PCD}$ with the weights $w,\varphi$ \;
Update the teacher model $T_\theta\leftarrow S_\theta$\;}
}
\Return parameters $\theta$ of student model {$S_\theta$}
\end{algorithm}
\end{sloppypar}